\documentclass[11pt]{article}

\usepackage[]{acl}

\usepackage{times, graphicx, footmisc, layouts}
\usepackage{latexsym,multirow}

\usepackage[utf8]{inputenc}

\usepackage{microtype}

\usepackage{times}
\usepackage{latexsym}
\usepackage{svg}
\usepackage{pdfpages}
\usepackage{float}
\usepackage{graphicx}
\usepackage{inconsolata}
\usepackage{booktabs}
\usepackage{lipsum}
\usepackage{graphicx}
\usepackage{amsmath}
\usepackage{adjustbox}
\usepackage{enumitem}
\usepackage{comment}

\renewcommand{\paragraph}[1]{\vspace{1mm}\noindent\textbf{#1}}

\newcommand{\bz}{\mathbf{z}}
\newcommand{\bv}{\mathbf{v}}
\newcommand{\bu}{\mathbf{u}}
\newcommand{\br}{\mathbf{r}}
\newcommand{\bh}{\mathbf{h}}
\newcommand{\bm}{\mathbf{m}}
\newcommand{\be}{\mathbf{e}}
\newcommand{\bk}{\mathbf{k}}
\newcommand{\bq}{\mathbf{q}}
\newcommand{\ba}{\mathbf{a}}
\newcommand{\brho}{\boldsymbol{\rho}}

\newcommand{\mcG}{\mathcal{G}}
\newcommand{\mcV}{\mathcal{V}}
\newcommand{\mcE}{\mathcal{E}}
\newcommand{\mcR}{\mathcal{R}}

\newcommand{\eg}{\textit{e}.\textit{g}.}
\newcommand{\roberta}{RoBERTa-large}

\newcommand{\methodname}{GrapeQA}

\makeatletter
\DeclareRobustCommand\onedot{\futurelet\@let@token\@onedot}
\def\@onedot{\ifx\@let@token.\else.\null\fi\xspace}

\makeatother

\title{GrapeQA: GRaph Augmentation and Pruning to Enhance Question-Answering}

\author{Dhaval Taunk$^*$, Lakshya Khanna$^*$, Pavan Kandru$^*$,\\ {\bf Vasudeva Varma, Charu Sharma and Makarand Tapaswi} \\
IIIT Hyderabad, India\\ \{dhaval.taunk,lakshya.khanna,siri.venkata\}@research.iiit.ac.in\\ \{vv,charu.sharma,makarand.tapaswi\}@iiit.ac.in}

\begin{document}
\maketitle
\begin{abstract}
Commonsense question-answering (QA) methods combine the power of pre-trained Language Models (LM) with the reasoning provided by Knowledge Graphs (KG).
A typical approach collects nodes relevant to the QA pair from a KG to form a Working Graph (WG) followed by reasoning using Graph Neural Networks (GNNs).
This faces two major challenges:
(i) it is difficult to capture all the information from the QA in the WG, and
(ii) the WG contains some irrelevant nodes from the KG.
To address these, we propose \methodname{} with two simple improvements on the WG:
(i) Prominent Entities for Graph Augmentation
identifies relevant text chunks from the QA pair and augments the WG with corresponding latent representations from the LM, and
(ii) Context-Aware Node Pruning
removes nodes that are less relevant to the QA pair.
We evaluate our results on OpenBookQA, CommonsenseQA and MedQA-USMLE and see that \methodname{} shows consistent improvements over its LM + KG predecessor (QA-GNN in particular) and large improvements on OpenBookQA.
\end{abstract}

\section{Introduction}
\def\thefootnote{*}\footnotetext{These authors contributed equally to this work.}
\def\thefootnote{\arabic{footnote}}



Answering questions is a challenging NLP problem as it involves understanding the question context and sifting through relevant information to identify the answer.
Question-answering models have evolved from rule-based~\cite{7980526} to RNN-based sequence models~\cite{https://doi.org/10.48550/arxiv.1703.05851} and now to Transformer-based Language Models (LM) such as \roberta~\cite{DBLP:journals/corr/abs-1907-11692}.
However, commonsense question-answering
adds a layer of complexity as the model needs to reason about questions relating diverse topics, making the task challenging for LMs that may not have seen something similar in the pre-training data.

While LMs capture the implicit patterns and contextual information within the data, KGs are able to capture explicit relations between the text entities.
KGs such as
Freebase~\cite{bollacker2008freebase},
Wikidata~\cite{vrandevcic2012wikidata}, or
ConceptNet~\cite{speer2017conceptnet} store knowledge in the form of graph triplets (topic-relationship-topic) and are well suited for Graph Neural Networks (GNNs), \eg ~\cite{welling2016semi}.
Thus, commonsense QA in particular has attracted interest in combining LMs and KGs with the reasoning ability of GNNs~\cite{lin-etal-2019-kagnet,yasunaga-etal-2021-qa}.


Most works on LM + KG extract a sub-graph or Working Graph (WG) from the KG based on concepts mentioned in the QA pair~\cite{lin-etal-2019-kagnet,feng-etal-2020-scalable,yasunaga-etal-2021-qa} and focus on improving reasoning.
For example, \citet{lin-etal-2019-kagnet}~propose a graph network to score answers while~\citet{feng-etal-2020-scalable} focus on a multi-hop message passing framework that allows each node to attend to multi-hop neighbors in a single layer, combining interpretable path-based reasoning with scalable GNNs.
\citet{yasunaga-etal-2021-qa}~improve the extracted WG through a relevance scoring mechanism followed by joint reasoning and~\citet{zhang2022greaselm} fuse information from both the modalities (LM, KG) by mixing their tokens and nodes.

Our emphasis with GrapeQA lies in improving the working graph (WG) with two simple ideas.
(i)~We augment the WG with useful information from the question-answer pair reducing the burden on a single QA context node used in previous works.(discussed in \ref{22})
(ii)~Instead of keeping all nodes of the WG, or simply scoring relevance, we drop less relevant information (nodes) from the WG simplifying the graph reasoning process.
The improvements to the WG are combined with the reasoning process of QA-GNN~\cite{yasunaga-etal-2021-qa} and evaluated on three datasets, where we see especially large improvements on domain-specific OpenBookQA (discussed in \ref{23}).




\section{GrapeQA Methodology}

\begin{figure*}[t!]
\centering 
\includegraphics[width=0.95\textwidth]{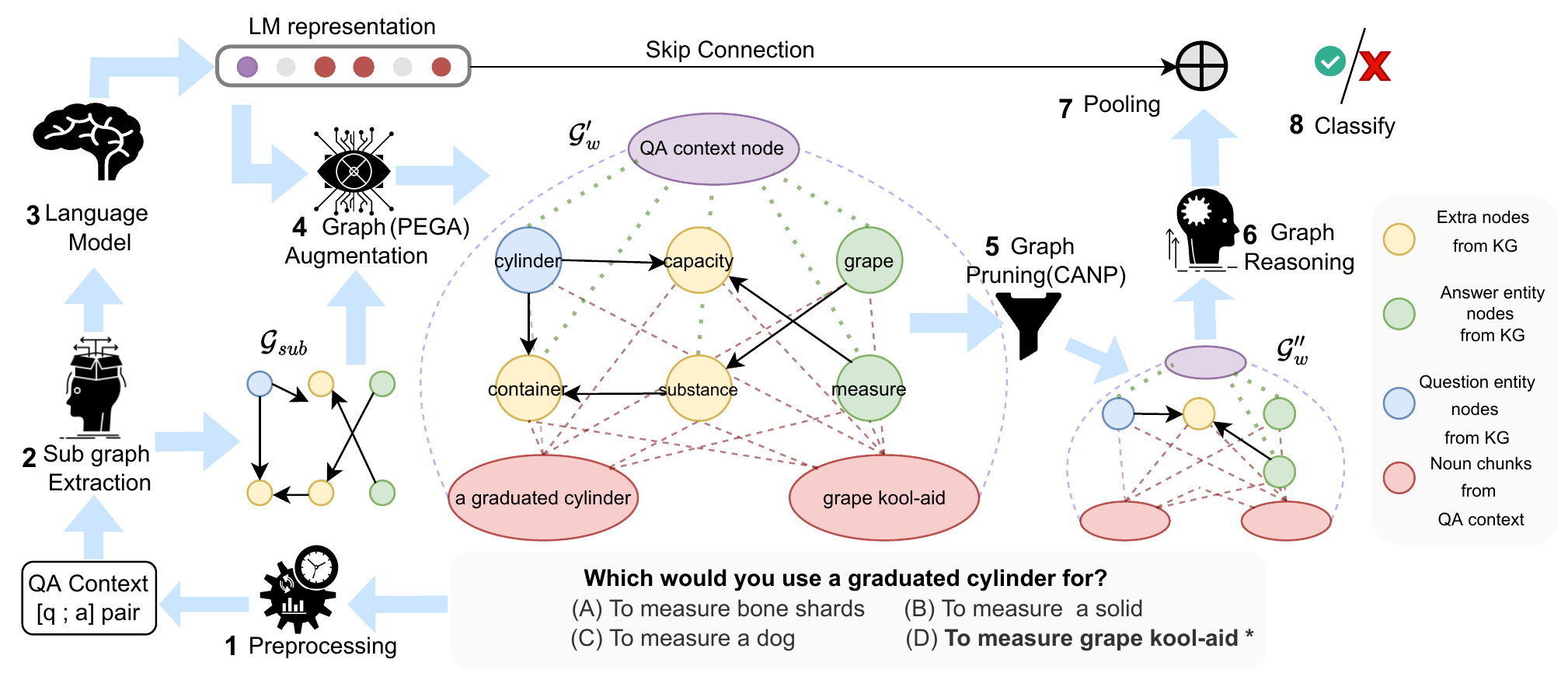}
\vspace{-3mm}
\caption{Method overview showing the approach to score the question with each answer option.
GrapeQA improves QA-GNN~\cite{yasunaga-etal-2021-qa} by augmenting the Working Graph with additional nodes that capture information from the QA pair (step 4: PEGA) and then pruning the graph to remove the least relevant nodes (step 5: CANP).}
\label{fig:model_overview}
\end{figure*}

We briefly describe the QA-GNN approach before our graph augmentation and pruning strategies.

\subsection{LM + KG: QA-GNN as a case study}
The objective of QA-GNN~\cite{yasunaga-etal-2021-qa} is to use both LM and KG for commonsense QA tasks.
Each multiple-choice QA consists of a question $q$ and $O$ answer options $\{a_o\}_{o=1}^O$ where only one is correct.
We create one Working Graph (WG) per answer option and reason over the graph to produce a score.
During training, cross-entropy loss is applied to scores of all answer options while we pick the highest scoring answer for inference.

We discuss the WG creation process starting with the KG.
Let $\mcG = (\mcV, \mcE)$ be the KG with $\mcV$ nodes and a set of edges $\mcE \subseteq \mcV \times \mcR \times \mcV$ with $\mcR$ relation types. 
For a given question-answer pair $[q; a_o]$, all nodes in the KG may not be relevant.
Hence, \emph{Question / Answer entity nodes}, referred as $q_\text{KG}$ or $a_\text{KG}$, that have some text matching with the question $q$ or answer option $a_o$ are picked.
Indirect relations between Question and Answer entity nodes are captured through common neighbors (2-hop away) by including them as \emph{Extra nodes} $s_\text{KG}$.
The sub-graph $\mcG_\text{sub}$ is formed together with the edges in $\mcE$ that connect the chosen KG nodes.
In summary, the nodes of the sub-graph are $\{q_\text{KG}\} \cup \{a_\text{KG}\} \cup \{s_\text{KG}\}$.

Next, a relevance scoring mechanism is used to prune irrelevant nodes that may appear in the sub-graph.
Scores are computed by encoding the \emph{QA context} (concatenated question and answer option text) and node label using an LM followed by a linear projection.
The relevance score influences the node representation in the sub-graph.
Finally, to create the Working Graph $\mcG_w$, \emph{QA context} is added as a node to the sub-graph and connected with other nodes using a new edge type.

Question, Answer, and Extra nodes in $\mcG_w$ are initialized by creating sentences based on triplets from the KG, feeding them to a pretrained LM, and average pooling over relevant tokens (see~\cite{feng-etal-2020-scalable} for details).
The QA context node is initialized as $\bz$, an encoding of the $[q; a_o]$ text using an LM.
To perform reasoning, a relation type aware Graph Network is adopted.
The output representations for all nodes are pooled and added to the LM's original encoding of the QA context.
Finally, an MLP is used to predict a score for the correctness of the answer option.
Fig.~\ref{fig:model_overview} illustrates QA-GNN along with our proposed modifications.
Additional details of QA-GNN are in App.~\ref{app:qagnn}.

\subsection{Graph Augmentation and Pruning}
\methodname{} proposes two improvements to the WG and corresponding adaptations to QA-GNN.
We overcome the limited capacity of the WG to exchange useful information between the QA context and the KG with \underline{P}rominent \underline{E}ntities for \underline{G}raph \underline{A}ugmentation (PEGA) that introduces additional nodes from the QA pair to the WG.
We also propose QA-\underline{C}ontext-\underline{A}ware \underline{N}ode \underline{P}runing (CANP), a pruning method that removes least relevant nodes.

\paragraph{Prominent Entities for Graph Augmentation (PEGA).}
\label{22}
Graph augmentation begins by extracting noun phrase chunks $c$ from the question and answer pair $[q; a_o]$.
We use Spacy's~\cite{Honnibal_spaCy_Industrial-strength_Natural_2020} \emph{noun} chunk extractor $f_\text{ext}$ to obtain
\begin{equation}
\mcV^{\prime}\ =\{c \mid c \in f_\text{ext}([q; a]) \} \, .
\end{equation}
The QA context is fed as input to the LM and representations of all the sub-word tokens are obtained.
Each extracted noun phrase is represented by averaging over the embeddings of its sub-word tokens.
As part of augmentation, these \emph{noun chunks nodes} ($\mcV^\prime$) are added as new nodes of type $n$ to the working graph $\mcG_{w}$.
\emph{Noun chunk nodes} also have two types of edges: $r_{no}$ between all the new ($\mcV^\prime$) and old $\mcG_w$ nodes, and $r_{nn}$ among the noun chunks themselves resulting in an augmented WG, $\mcG_w^\prime$:
\begin{eqnarray}
&\mcE^{\prime}\ = \{\mcV^\prime \times r_{nn} \times \mcV^\prime\} \cup \{\mcV^\prime \times r_{no} \times \mcV\} \, , \\
&\mcG_{w}^{\prime} = (\mcV \cup \mcV^{\prime},  \mcE \cup \mcE^{\prime}) \, .
\end{eqnarray}


\paragraph{QA Context-Aware Node Pruning (CANP)}
\label{23}
aims to remove the less relevant nodes from the WG.
Our intuition is that some Extra nodes (i.e.~2-hop neighbors from the KG which do not match the QA text) may be less relevant to the QA as compared to the Question / Answer entity nodes.

To perform pruning, we first associate and cluster Extra nodes with \emph{Answer entity nodes}.
CANP is only applied when there are more than one Answer entity nodes.
Recall that the WG is created for one answer option (or one QA pair) and the number of Answer entity nodes (and clusters) depends on the number of nodes with text similar to the answer option in the KG.
Similar to relevance scoring in QA-GNN, we calculate the relevance score for each Extra node $s_\text{KG}$ against each Answer entity $a_\text{KG}$ by encoding the concatenated text of the QA pair, the Answer entity, and the Extra node.
\begin{equation}
\psi_{sa}^\text{KG} = f_{\text{head}} \left( \text{LM} \left( [\text{text}(\bz); \text{text}(\ba_\text{KG}); \text{text}(\bs)] \right) \right) \, ,
\end{equation}
where $\text{text}(\cdot)$ corresponds to the node's label text:
$[q; a_o]$ pair for $\bz$,
Extra node's label $s_\text{KG}$ for $\bs$, and
the Answer entity label $a_\text{KG}$ for $\ba_\text{KG}$.
Thus, each Extra node $s_\text{KG}$ is assigned to the cluster $\mcV_x$ corresponding to the highest relevance score,
\begin{equation}
\mcV_{x} = \{s_\text{KG} \, | \, x = \arg\,\max_{a_\text{KG}} \, \psi_{sa}^\text{KG} \} \, .
\end{equation}
We compute the average relevance score for each cluster and identify the least relevant cluster $\mcV_r$ as
\begin{eqnarray}
\psi_x^\text{KG} &=& \sum_{s_\text{KG} \in \mcV_x} \psi_{sx}^\text{KG} / |\mcV_x| \, , \\
\mcV_{r} &=& \mcV_{x} \,\, \text{ s.t. } \, r = \arg\,\min_x \psi_x^\text{KG} \, .
\end{eqnarray}
Finally, we remove the cluster with lowest average relevance score from the WG before continuing with graph-based reasoning.
The PEGA augmented WG can be pruned as
\begin{equation}
\mcG_{w}^{\prime\prime} = (\mcV \cup \mcV^{\prime} -\mcV_{r} , \mcE \cup \mcE^{\prime} - \{\mcV_{r} \times R \times \mcV\}).
\end{equation}

\section{Experiments}
\label{sec:experiments}

\begin{table}[b]
\caption{Average number of nodes of each type in WGs.}
\centering
\tabcolsep=0.12cm
\begin{tabular}{lccc}
\toprule
\textbf{Node Type} & \textbf{OBQA}  & \textbf{CSQA}  & \textbf{MedQA}  \\
\midrule
Question entity $q_\text{KG}$  & 6.52          & 7.36          & 6.1             \\
Answer entity $a_\text{KG}$    & 2.79          & 2.05          & 0.55            \\
Extra nodes $s_\text{KG}$    & 107.17        & 112.04        & 20.82           \\
Noun chunk nodes $\mcV^\prime$ & 3.88          & 4.13          & 33.46          \\
\bottomrule

\end{tabular}

\label{tab:nodecount}
\end{table}




\begin{table*}[!t]
\caption{Comparison of Accuracy between LM+KG methods on the OpenBookQA, CommonsenseQA (left) and MedQA (right).}
\centering
\tabcolsep=0.12cm
\begin{adjustbox}{max width=\textwidth}
\begin{tabular}{lcc}
\toprule
& \bf OBQA & \bf CSQA \\
\textbf{Model} & \textbf{Test}  & \textbf{IHTest} \\
\midrule
RGCN~\small{\cite{schlichtkrull2018modeling}} & 62.45 & 68.4\\
GconAttn~\small{\cite{wang2019improving}} & 64.75 & 68.6\\
RN~\small{\cite{santoro2017simple}} & 65.20 & 69.1\\
MHGRN~\small{\cite{feng-etal-2020-scalable}} & 66.85 & 71.1\\
GreaseLM (AristoRoBERTa)~\small{\cite{zhang2022greaselm}} & \underline{84.8} & \underline{74.05}\\
\midrule
QA-GNN (\roberta)~\small{\cite{yasunaga-etal-2021-qa}} & 67.80 & 73.4 \\
\methodname{}: CANP (\textbf{Ours}) & 66.20 & \bf 74.94\\
\methodname{}: PEGA (\bf \textbf{Ours}) & 82.0 & 73.41 \\
\methodname{}: PEGA+CANP (\textbf{Ours}) & \bf 90.0 & 74.05 \\
\bottomrule
\end{tabular}
\quad
\begin{tabular}{lc}

\toprule
\multicolumn{2}{c}{\bf MedQA} \\
\textbf{Model} & \textbf{Test}\\
\midrule
BERT-base~\small{\cite{devlin-etal-2019-bert}} & 34.3 \\
BioBERT-base~\small{\cite{10.1093/bioinformatics/btz682}} & 34.1 \\
\roberta~\small{\cite{DBLP:journals/corr/abs-1907-11692}} & 35.0 \\
BioBERT-large~\small{\cite{10.1093/bioinformatics/btz682}} & 36.7 \\
SapBERT~\small{\cite{https://doi.org/10.48550/arxiv.2010.11784}} & 37.2 \\
GreaseLM~\small{\cite{zhang2022greaselm}} & \underline{38.5} \\
\midrule
QA-GNN~\small{\cite{yasunaga-etal-2021-qa}} & 38.0 \\ 
\methodname{}: (PEGA) (\bf \textbf{Ours}) & \bf 39.51 \\
& \\
\bottomrule

\end{tabular}
\end{adjustbox}

\label{tab:combined}
\end{table*}




\begin{table}[t]
\caption{PEGA Ablations: Impact of different noun chunk extraction methods on OBQA.}
\centering
\vspace{1mm}
\begin{tabular}{cc}
\toprule
\textbf{Noun chunk extraction method} & \textbf{Accuracy} \\
\midrule
20\% random words & 72.32\\ 
NLTK~\small{\cite{journals/corr/cs-CL-0205028}} & 78.40 \\ 
spaCy~\small{\cite{Honnibal_spaCy_Industrial-strength_Natural_2020}} & \bf 82.00 \\ 
\bottomrule
\end{tabular}

\label{tab:noun_chunk_ablations}
\end{table}

We evaluate \methodname{} on three QA datasets: \\
\textbf{1.~CommonsenseQA} (CSQA) is 5-way multiple-choice QA (MC-QA) dataset of 12,102 questions that requires commonsense reasoning to answer questions.
We use standard splits~\cite{lin-etal-2019-kagnet} and report results on the in-house test (IHtest).
\textbf{2.~OpenBookQA} (OBQA) is a 4-way MC-QA dataset of 5,957 questions based on elementary science knowledge; splits by~\citet{mihaylov2018can}.
\textbf{3.~MedQA-USMLE} is a 4-way MC-QA dataset based on biomedical and clinical knowledge and has 12,723 questions from United States Medical License Exams, with splits by~\citet{jin2021disease}.

Table~\ref{tab:nodecount} presents node counts in the WG for the above datasets while Table~\ref{tab:overlap} (App.~\ref{app:graphstats}) shows a small overlap between noun chunk and KG nodes.


\paragraph{Implementation \& training details.}
The LM adopted in our work is \roberta~\cite{DBLP:journals/corr/abs-1907-11692} for CSQA and OBQA, and SapBERT~\cite{https://doi.org/10.48550/arxiv.2010.11784} for MedQA.
ConceptNet~\cite{speer2017conceptnet} is our KG for generating the WG in CSQA and OBQA.
For MedQA, we use the graph constructed by QA-GNN~\cite{yasunaga-etal-2021-qa}.
%
Our model consists of an LM and a GNN with dim 200.
RADAM optimizer is used with a learning rate of $10^{-5}$ for the LM and $10^{-3}$ for the GNN.
OBQA \& MedQA are trained for 50 epochs with a batch size of 128 and CSQA for 20 epochs with a batch size of 64.
All models are a single run trained on 2 RTX 2080 Ti GPUs and take about 28 hours for OBQA and 16 hours for CSQA and MedQA.

\subsection{Comparisons with Baselines}
We use accuracy as a metric and compare our results primarily against other works that also adopt LM + KG methods (see Table \ref{tab:combined}).
\methodname{} builds on top of QA-GNN (for direct comparison) and improving the WG results in highest performance on OBQA \& MedQA and comparable performance on CSQA.
For a fair comparison, we use the same LM for all methods unless noted.

LM only methods tend to perform worse than the baseline QA-GNN.
\roberta~\cite{DBLP:journals/corr/abs-1907-11692} for CSQA provides 72.1\% while \roberta and AristoRoBERTa~\cite{clark2019aristoroberta} for OBQA show 64.80\% and 77.8\%, respectively.
For MedQA, the LM only model results are also shown in Table~\ref{tab:combined} (right); we see that LMs trained on medical data (\eg~SapBERT~\cite{https://doi.org/10.48550/arxiv.2010.11784}) outperform generic LMs on this domain-specific task.
\methodname{} outperforms all these approaches.



\paragraph{OBQA.}
CANP applied to the original QA-GNN WG is unable to improve performance (-1.6\%), probably because the WG is not rich.
However, PEGA provides a 14.2\% accuracy improvement over QA-GNN (82\% vs. 67.8\%).
Interestingly, CANP when used together with PEGA boosts the accuracy to 90\% (+22.2\%);
surpassing GreaseLM that uses an improved LM (AristoRoBERTa) and better integration of LM + KG by 5.2\%.
For the \emph{domain-specific} OBQA, PEGA adds relevant information while CANP effectively cleans up irrelevant nodes resulting in large improvements.


\paragraph{MedQA.}
PEGA achieves an improvement of 1.5\% over QA-GNN, and 1\% over GreaseLM, the previous SoTA.
A reason for the small improvement (compared to OBQA) could be that the WG for MedQA has fewer nodes (see Table~\ref{tab:nodecount}).
Additionally, the small number of Answer entity nodes in the WG also means that CANP is not applicable.

\paragraph{CSQA.}
On \emph{generic commonsense} questions, the WG can have large amounts of irrelevant information that CANP can simplify.
We see an improvement of 1.5\% over QA-GNN when using CANP only.
However, unlike OBQA, PEGA shows comparable performance to QA-GNN as it may lead to stuffing the WG with common terms (noun chunks) that do not provide discriminatory information.
Nevertheless, CANP alone also improves over GreaseLM by 0.9\% (all in absolute points).

\subsection{Ablation experiments}
\label{sec:ablation}



\paragraph{Noun chunk extraction.}
While PEGA is an effective graph augmentation strategy, it relies on the noun chunk extraction method.
We evaluate automatic noun chunk extraction methods spaCy and NLTK (see App.~\ref{app:spacy_nltk} for details) against a simple baseline that randomly adds 20\% of the QA pair's words to the WG.
Table~\ref{tab:noun_chunk_ablations} shows that extracting meaningful chunks is important and may lead to large performance change (on OBQA).
Interestingly, even random chunks of the QA pair provides a 4.5\% boost over QA-GNN that only includes one node to encode the entire QA context.

\paragraph{Number of GNN layers}
is often an important hyperparameter.
We show results for both the PEGA+CANP (Table~\ref{tab:num_layers_pc}) and PEGA-only (Table~\ref{tab:num_layers_p}) models in Appendix~\ref{app:more_results}.
Generally, 5 layers seem to work well for all settings, while the CSQA PEGA-only model shows better results with 4 layers.

\section{Conclusion}
\label{sec:conclusion}

We presented GrapeQA, an effective approach to integrate information from QA (LM) and KG for commonsense QA.
We proposed two simple improvements to the working graph:
PEGA, a graph augmentation that improves information flow between the QA and the KG; and
CANP that prunes less relevant information.
Our approach led to new SoTA results on three datasets OBQA, CSQA, and MedQA, with a large 22\% increase on OBQA.

\section{Ethical Impact}
\label{sec:ethical_impact}

In order to support commonsense thinking, this study suggests a general method for fusing language models and external knowledge graphs.
We rely on publicly available datasets and benchmarks and knowledge graphs for each experiment.
We could not anticipate any immediate social ramifications or ethical concerns as we neither amplify existing bias in the data nor do we inject any social or ethical bias into the model.

\bibliography{main}
\bibliographystyle{acl_natbib}

\appendix
\clearpage
\begin{center}
\textbf{\Large{Appendix}}
\end{center}

We first present additional results in Appendix~\ref{app:more_results} followed by a detailed explanation of QA-GNN in Appendix~\ref{app:qagnn}.
Appendix~\ref{app:graphstats} provides some statistics for the datasets and working graphs, while Appendix~\ref{app:spacy_nltk} presents details of noun chunk extraction methods used in PEGA.

\section{Additional Results}
\label{app:more_results}

\paragraph{Number of GNN layers ablations.}
Tables~\ref{tab:num_layers_pc} and \ref{tab:num_layers_p} show ablation studies by varying the number of GNN layers over PEGA+CANP and PEGA-only respectively.
5 layer GNNs seem to be a suitable for both methods, while CSQA with PEGA-only shows highest performance with 4 layers.

\paragraph{CANP is not necessary on MedQA.}
Table~\ref{tab:nodecount} of the main paper shows the average number of nodes of different types in a WG.
The number of extra concept nodes is much higher than the QA concept nodes except in the MedQA dataset.
This makes it necessary to prune these nodes to keep only the relevant ones.
In case of MedQA since the number of extra nodes in WG are already quite low, and the nodes from the KG are often meaningful (domain-specific) we do not perform CANP pruning.

\paragraph{Results on CSQA.}
Table~\ref{tab:csqa_off} shows the results of our model on the official test set for CommonsenseQA.
We compare our results with other existing approaches, both using powerful LMs (e.g.,~UnifiedQA) or LM+KG methods (QA-GNN, GreaseLM, etc.).
Unfortunately we were unable to evaluate our best performing model on the in-house test set (GrapeQA: CANP-only) due to limited number of submissions indicated for the evaluation.
Even on the in-house test set, we see no performance change between PEGA-only and QA-GNN (73.41\% vs. 73.4\%) while a $\pm$1\% variation exists due to random seeds.

\begin{table}[h]
\caption{Impact of the number of GNN layers using the PEGA+CANP model.}
\centering
\tabcolsep=0.10cm
\begin{tabular}{lcc}
\toprule
 & \multicolumn{2}{c}{ \textbf{Accuracy} } \\
\textbf{\#layers} & \textbf{OBQA} & \textbf{CSQA} \\
\midrule
4 & 88.38 &  72.60\\ 
5 & \bf 90.00 &  \bf 74.05\\
6 & 88.96 & 71.88\\
\bottomrule

\end{tabular}

\label{tab:num_layers_pc}
\vspace{5mm}
\caption{Impact of the number of GNN layers using the PEGA only model.}
\centering
\begin{tabular}{lcc}

\toprule
 & \multicolumn{2}{c}{ \textbf{Accuracy} } \\
\textbf{\#layers} & \textbf{OBQA} & \textbf{CSQA} \\
\midrule
4 & 83.20 & \bf 74.62\\ 
5 & \bf 82.00 &  73.41\\
6 & 81.40 & 73.17\\
\bottomrule
\end{tabular}

\label{tab:num_layers_p}
\end{table}

\begin{table}[t]
\small
\centering
\tabcolsep=0.02cm
\caption{Comparison on CommonSenseQA official test set using \roberta model. The best result is in \textbf{bold} and second best is \underline{underlined}.
Due to limited entries for evaluation, we were unable to evaluate our best method on CSQA: CANP-only.
*UnifiedQA has 11B parameters and is about 30x larger than QA-GNN and our model and is trained on much more data.}
\begin{tabular}{lc}

\toprule
\textbf{Model} & \textbf{Test Acc.}\\
\midrule
RoBERTa~\small{\cite{DBLP:journals/corr/abs-1907-11692}} & 72.1 \\
RoBERTa + FreeLB (ensemble)~\small{\cite{https://doi.org/10.48550/arxiv.1909.11764}}& 73.1 \\
RoBERTa + HyKAS~\small{\cite{https://doi.org/10.48550/arxiv.1910.14087}} & 73.2 \\
RoBERTa + KE (ensemble) & 73.3 \\
RoBERTa+KEDGN (ensemble) & 74.4 \\
XLNet+GraphReason~\small{\cite{https://doi.org/10.48550/arxiv.1909.05311}} & 75.3 \\
RoBERTa+MHGRN~\small{\cite{feng-etal-2020-scalable}} & 75.4 \\
Albert+PG~\small{\cite{wang-etal-2020-connecting}} & 75.6 \\ 
QA-GNN \small{\cite{mihaylov2018can}} & 76.1 \\
Albert (ensemble)~\small{\cite{conf/iclr/LanCGGSS20}} & 76.5 \\
UnifiedQA*~\small{\cite{https://doi.org/10.48550/arxiv.2005.00700}} & 79.1 \\
\midrule
\methodname{} (PEGA) (\bf \textbf{Ours}) & \bf 73.5 \\
\bottomrule
\end{tabular}

\label{tab:csqa_off}
\end{table}

\section{QA-GNN Method Details}
\label{app:qagnn}

We provide further details of the question-answering procedure adopted by QA-GNN~\cite{yasunaga-etal-2021-qa}.

\subsection{Relevance Scoring}
\label{sec_app:relevance_scoring}
Extracting a sub-graph by selecting few hop neighbors adds many irrelevant nodes to the sub-graph.
QA-GNN proposes a relevance scoring mechanism to add an ``importance score" to the initial embedding of concept nodes. This helps the GNN to focus on nodes with high relevance score while performing graph reasoning.
\begin{align}
\brho_{v} &= f_{\text{head}}\left(f_{\text{enc }}([\operatorname{text}(\bz) ; \operatorname{text}(\bv)])\right) \, , \\
\rho_{t} &=f_{\rho}\left(\brho_{v}\right) \, .
\end{align}
The text of each concept node $\text{text}(\bv)$ is concatenated with the QA-pair (referred to as $\text{text}(\bz)$).
The LM encoding and following an MLP head produces an embedding $\brho_v$ that is converted into a relevance score $\rho_{v}$ using an MLP $f_{\rho}$.
Nodes with a score lower than a threshold are discarded.

\subsection{Node and Relation Types}
QA-GNN constructs a working graph which is heterogeneous and multi-relational.
It uses a node type ($u$) aware and relation type ($r$) aware iterative message passing network to reason over it.
Different node types are represented using embeddings $\bu$. These include QA context, Question entity, Answer entity and Extra nodes. 
Whereas, edge embeddings $\be$ include relations in KG and two new relation types between the QA context node and KG entity nodes.

Node and relation types are embedded using MLPs $f_{u}$ and $f_{r}$ respectively,
\begin{align}
\bu_{t} &= f_{u} \left( u_{t} \right) \, , \\
\br_{s t} &= f_{r} \left( \be_{s t}, u_{s}, u_{t} \right) \, .
\end{align}
The message from the source to target node is constructed by concatenating the source node and type representations along with the relation embedding from the source to target and projecting it (to node embedding dimension) using the MLP $f_{m}$. 
\begin{equation}
\bm_{s t} = f_{m} \left( \bh_{s}^{(\ell)}, \bu_{s}, \br_{s t} \right) \, .
\end{equation}

\subsection{Message Passing}
Node representations are updated at each layer using the following attention mechanism
\begin{equation}
\bq_{s} = f_{q} \left( \bh_{s}^{(\ell)}, \bu_{s}, \brho_{s} \right) \, ,
\end{equation}
and
\begin{equation}
\bk_{t} = f_{k} \left( \bh_{t}^{(\ell)}, \bu_{t}, \brho_{t}, \br_{s t} \right) \, .
\end{equation}

The query $\bq_{s}$ and key $\bk_{t}$ vectors of the source and target nodes are computed using the node representation $\bh$, the node type embedding $\bu$ and relevance score embeddings $\brho$.
Finally, we score attention $\alpha_{st}$ as
\begin{equation}
\alpha_{s t}=\frac{\exp \left(\gamma_{s t}\right)}{\sum_{t^{\prime} \in \mathcal{N}_{s} \cup\{s\}} \exp \left(\gamma_{s t^{\prime}}\right)}, \,\, \gamma_{s t}=\frac{\boldsymbol{q}_{s}^{\top} \boldsymbol{k}_{t}}{\sqrt{D}} \, .
\end{equation}

The attention weights for messages from source to target $\alpha_{st}$ are calculated using $\bq$ and $\bk$ vectors.
Finally, we aggregate messages and update the node representation as
\begin{equation}
\label{eq:17}
\boldsymbol{h}_{t}^{(\ell+1)}=f_{n}\left(\sum_{s \in \mathcal{N}_{t} \cup\{t\}} \alpha_{s t} \boldsymbol{m}_{s t}\right)+\boldsymbol{h}_{t}^{(\ell)} \, .
\end{equation}

\begin{table}[t]
\caption{Number of unique nodes across all WG of the dataset.
Even though more nodes are added from the KG on average (see Table~\ref{tab:nodecount}), they are not all unique across the dataset and result in a smaller count.}
\centering
\tabcolsep=0.12cm
\begin{tabular}{l ccc}
\toprule
    & \textbf{Noun chunk} & \textbf{Nodes}    & \textbf{Overlapping} \\
\textbf{Dataset} & \textbf{nodes}  & \textbf{from KG}  & \textbf{nodes} \\
\midrule
OBQA  & 14470    & 7506  & 1958    \\
CSQA  & 23881    & 12485 & 4023    \\
MedQA & 69370    & 2753  & 1268    \\
\bottomrule
\end{tabular}

\label{tab:overlap}
\end{table}

\begin{table}[t]
\caption{Average number of words in the question $q$ and answer option $a_o$ for the different datasets.}
\centering
\tabcolsep=0.12cm
\begin{tabular}{l cc}
\toprule
  & \textbf{Question} & \textbf{Answer} \\
\midrule
OBQA  & 13.5     & 2.8 \\
CSQA  & 13.8     & 1.5 \\
MedQA & 116.2    & 3.6 \\
\bottomrule
\end{tabular}

\label{tab:wordcount}
\end{table}

\section{Working Graph Statistics}
\label{app:graphstats}
Given a question and corresponding answer option, KG nodes with matching text entities are identified.
These matched nodes along with the Extra nodes that fall in 2-hop paths from them form the sub-graphs for each $[q; a]$ pair.
Working Graphs are constructed by joining these sub-graphs with QA context nodes initialized with the representation from LM.
In each Working Graph, the QA context node is connected to all the concept nodes in it which are extracted from the KG. 

\paragraph{Node counts.}
Table~\ref{tab:nodecount} in the main paper shows the number of nodes added to the WG on average.
We see that general KGs (ConceptNet) afford a large number of extra nodes (100+) while MedQA with a smaller KG only adds a few extra nodes ($\sim$20).
The large number of noun chunks added in the MedQA is explained by the fact that the questions in MedQA are very large as they include patient's description.
Table~\ref{tab:wordcount} presents the average number of words in the question and answer option.

\paragraph{Noun chunks are unique.}
Table~\ref{tab:overlap} shows the number of unique nodes present in each dataset.
It can be observed that the total number of \emph{unique} nodes selected from the KG is low as compared to the total number of unique noun chunk nodes extracted.
Even though Table~\ref{tab:nodecount} shows that a large number of nodes are added to the graph, they are not all unique.
Thus, even if the average number of noun chunk nodes for each WG are low, they are more diverse compared to nodes from KG.
A small overlap between noun chunk nodes and nodes from the KG indicates that this way of constructing the WG may provide better opportunity for graph reasoning to exchange information effectively between the QA (LM) and the KG.

\section{Noun chunk extraction methods}
\label{app:spacy_nltk}

\paragraph{SpaCy}
is a Python and Cython programming language-based open-source software library for sophisticated natural language processing%
\footnote{\url{https://spacy.io/}}.
The library is distributed under the MIT licence.

In our experiments, we used \texttt{en\_core\_web\_sm} package which provides functionalities like tok2vec, tagger, parser, attribute\_ruler, lemmatizer, ner. We have used the noun chunk parser technique for extracting the noun chunks.

\paragraph{NLTK}
In order to work with human language data, Python programs can be built using the NLTK framework.
NLTK offers simple access to more than 50 corpora and lexical resources, including WordNet, as well as a number of text processing libraries for categorization, tokenization, stemming, tagging, parsing, and semantic reasoning.

In our implementation, we first tokenize the input sentence using NLTK's \texttt{word\_tokenizer}.
Then to extract the noun chunks, the POS tagger of NLTK is used.
\end{document}